\documentclass[journal]{IEEEtran}
\usepackage{graphicx}
\usepackage{flushend}
\usepackage{amsmath}
\usepackage{amssymb}
\usepackage{array}
\usepackage{booktabs}
\usepackage{listings}
\usepackage{tikz}
\usetikzlibrary{shapes.geometric, arrows.meta, positioning}
\usepackage{algorithm}
\usepackage{algpseudocode}
\usepackage{cite}
\usepackage{url}
\usepackage{tabularx}
\usepackage{caption}

\IEEEoverridecommandlockouts

\begin{document}

\title{Comparative Analysis of YOLOv5, Faster R-CNN, SSD, and RetinaNet for Motorbike Detection in Kigali’s Autonomous Driving Context}

\author{%
\IEEEauthorblockN{Ngeyen Yinkfu, Sunday Nwovu, Jonathan Kayizzi, Angelique Uwamahoro}\\
\IEEEauthorblockA{Carnegie Mellon University Africa, Kigali, Rwanda\\
\{nyinkfu, snwovu, jkayizzi, auwamaho\}@andrew.cmu.edu}
}

\maketitle

\begin{abstract}
In Kigali, Rwanda, motorcycle taxis are a primary mode of transportation, often navigating unpredictably and disregarding traffic rules, posing significant challenges for autonomous driving systems. This study compares four object detection models—YOLOv5, Faster R-CNN, SSD, and RetinaNet—for motorbike detection using a custom dataset of 198 images collected in Kigali. Implemented in PyTorch with transfer learning, the models were evaluated for accuracy, localization, and inference speed to assess their suitability for real-time navigation in resource-constrained settings. We identify implementation challenges, including dataset limitations and model complexities, and recommend simplified architectures for future work to enhance accessibility for autonomous systems in developing countries like Rwanda.
\end{abstract}


\section{Introduction}
In developing countries like Rwanda, motorcycle taxis, locally known as "moto taxis," dominate urban transportation, particularly in Kigali. These vehicles frequently disregard traffic regulations, weaving through traffic and creating dynamic, unpredictable environments for autonomous vehicles and robots. Robust object detection is critical to ensure safe navigation in such conditions. This study evaluates four state-of-the-art models—YOLOv5 \cite{jocher2020yolov5}, Faster R-CNN \cite{ren2015faster}, SSD \cite{liu2016ssd}, and RetinaNet \cite{lin2017focal}—for motorbike detection using a custom dataset of 198 images collected in Kigali, annotated in COCO format. Implemented in PyTorch with transfer learning, the models are compared to identify optimal solutions for autonomous driving in resource-constrained African cities.

Our contributions are:
\begin{enumerate}
    \item A comparative analysis of YOLOv5, Faster R-CNN, SSD, and RetinaNet for motorbike detection in Kigali’s urban context.
    \item Identification of challenges, including dataset limitations and model-specific complexities, relevant to developing countries.
    \item Recommendations for simplified architectures to enhance accessibility for autonomous systems in resource-limited settings.
\end{enumerate}

\section{Related Work}
Autonomous driving in Africa faces unique challenges due to the prevalence of motorcycles, particularly in urban centers like Kigali. Studies like \cite{adekunle2018autonomous} highlight the need for robust detection systems to handle non-compliant motorcycle taxis, which often navigate unpredictably. Research in Nairobi \cite{mutahi2020urban} underscores the complexity of detecting two-wheeled vehicles in congested, mixed-traffic environments. Existing work on African road safety \cite{okoth2021traffic} notes that motorcycles contribute significantly to traffic incidents, necessitating precise detection for autonomous navigation.

Object detection models are categorized as two-stage (e.g., Faster R-CNN \cite{ren2015faster}) or single-stage (e.g., YOLO \cite{redmon2016yolo}, SSD \cite{liu2016ssd}, RetinaNet \cite{lin2017focal}). Faster R-CNN uses a ResNet backbone and region proposal network (RPN) for high accuracy \cite{ren2015faster}. YOLOv5 employs CSPDarknet53 and FPN-PAN for balanced speed and accuracy \cite{jocher2020yolov5}. SSD leverages VGG16 for rapid detection \cite{liu2016ssd}, while RetinaNet introduces Focal Loss to address class imbalance \cite{lin2017focal}. Custom implementations often face challenges with complex decoding and loss functions, particularly in resource-constrained settings where computational infrastructure is limited.

\section{Literature Review}
The deployment of Advanced Driver Assistance Systems (ADAS) and autonomous vehicles (AVs) in Low- and Middle-Income Countries (LMICs) like Rwanda presents unique challenges, particularly in Kigali’s unstructured urban traffic dominated by motorcycle taxis. This section expands on the contextual imperatives, architectural trade-offs, and technical challenges of motorbike detection, focusing on small object detection (SOD), occlusion, and edge deployment.

\subsection{Contextual Imperative: Motorbike Detection in Unstructured Traffic}
Kigali’s traffic is characterized by high heterogeneity and unpredictability, with mixed flows of motorbikes, pedestrians, and vehicles navigating roads often lacking standardized signage or lane markings \cite{driveindia2025}. Unlike Western benchmarks (e.g., KITTI, nuScenes), Kigali’s environment requires detection algorithms to handle diverse object categories under variable conditions \cite{matec2024}. Motorbikes, as vulnerable road users, pose a safety-critical challenge due to their small size, rapid movement, and frequent occlusion \cite{motorcycle2017}. Accurate detection and localization are essential for real-time traffic monitoring and congestion management \cite{mdpi2024vehicle}. Rwanda’s pioneering Automated Speed Enforcement (ASE) system, scaled nationwide in 2021, has reduced crash-related deaths by decreasing mean speeds \cite{tirf2025, pubmed2025}, providing a policy foundation for advanced computer vision systems \cite{jepa2025}.

\subsection{Architectural Trade-offs in Object Detection}
Object detection models balance accuracy and inference speed, critical for ADAS in Kigali’s dynamic traffic.

\subsubsection{Two-Stage Detectors: Faster R-CNN}
Faster R-CNN’s two-stage process, with a Region Proposal Network (RPN) and subsequent classification, excels in localization accuracy (mAP@0.5:0.95) for dense scenes \cite{ren2015faster, scitepress2024}. However, its computational complexity limits real-time applicability unless optimized with lightweight backbones (e.g., Res2Net-101) or quantization \cite{carranza2021onstage, wang2021lightweight}.

\subsubsection{Single-Stage Detectors}
Single-stage detectors prioritize speed. YOLOv5’s CSPDarknet backbone, FPN-PAN neck, and Mosaic augmentation offer a robust speed-accuracy balance, with flexible model sizes (e.g., YOLOv5s) suitable for edge deployment \cite{jocher2020yolov5, ultralytics2025yolov5}. SSD’s VGG16 backbone enables fast inference but struggles with small objects due to limited semantic depth in shallow layers \cite{liu2016ssd, arxiv2017fssd}. Feature Fusion SSD (FSSD) mitigates this by enhancing feature maps \cite{arxiv2017fssd}. RetinaNet’s Focal Loss addresses class imbalance, making it effective for sparse, occluded motorbikes, with FPN ensuring multi-scale detection \cite{lin2017focal}.

\subsection{Challenges: Small Object Detection and Occlusion}
Motorbike detection in Kigali requires addressing small object detection (SOD) and occlusion. SOD is challenging due to low pixel coverage, degrading features during down-sampling \cite{arxiv2025smallobject}. Enhanced FPNs and attention mechanisms (e.g., channel and spatial attention in AFYOLO) improve feature extraction for small objects \cite{researchgate2021small}. Occlusion, common in dense traffic, is mitigated by contextual feature fusion and motion cues from optical flow in dual-stream networks \cite{researchgate2023occluded, arxiv2025uav}.

\subsection{Edge Deployment and Optimization}
Real-time ADAS in Kigali demands lightweight models for edge devices like NVIDIA Jetson. Quantization to 8-bit integers reduces memory and latency, enabling SSD MobileNetV2 to achieve low latency on Jetson Orin Nano \cite{beei2025}. Pruning and knowledge distillation further optimize models \cite{mdpi2025collaborative}. YOLOv5’s ecosystem supports rapid edge deployment, while Faster R-CNN and RetinaNet require aggressive optimization to meet real-time FPS thresholds \cite{ultralytics2025yolov5, fan2024lightweight}.

\section{Methodology}
\begin{figure}[ht]
    \centering
    \includegraphics[width=0.9\linewidth]{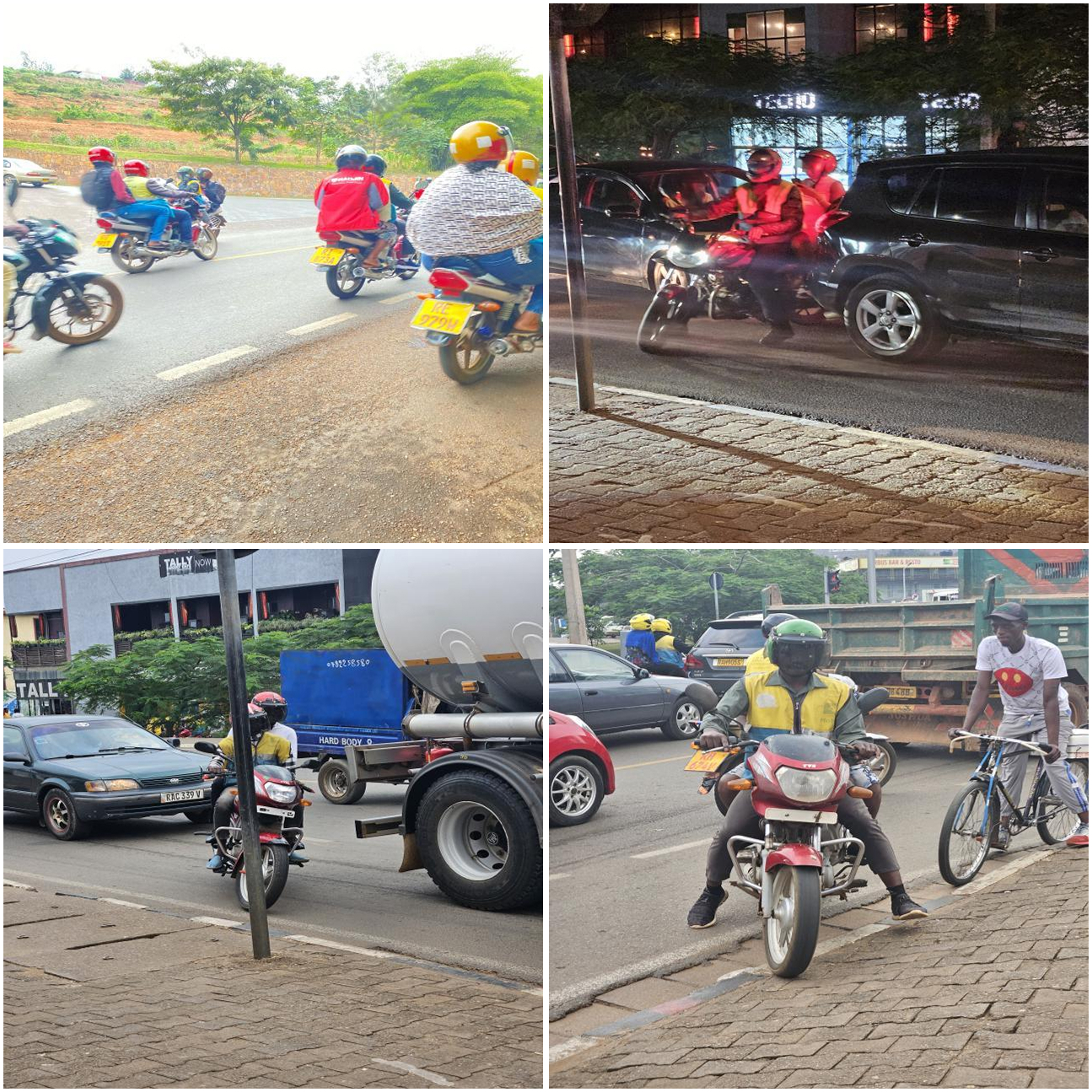}
    \caption{Example images of motorcycles taken in Kigali, both in the day and at night, and congested traffic scenarios.}
    \label{fig:sample_img}
\end{figure}
\subsection{Dataset}
A dataset of 198 motorbike images was collected in Kigali, capturing scenarios like congested traffic, night conditions, and non-compliant motorcycle taxis. Images were annotated in COCO format via Roboflow and split into:
\begin{itemize}
    \item Train: 168 images (244 annotations)
    \item Validation: 20 images (48 annotations)
    \item Test: 10 images (30 annotations)
\end{itemize}
Preprocessing involved JSON-to-DataFrame conversion, bounding box validation (excluding three invalid images), and orientation standardization (rotating 13 landscape images). The small dataset size limits generalizability, as discussed in Section \ref{sec:discussion}.

\subsection{Data Augmentation}
\begin{figure}[h]
    \centering
    \includegraphics[width=1\linewidth]{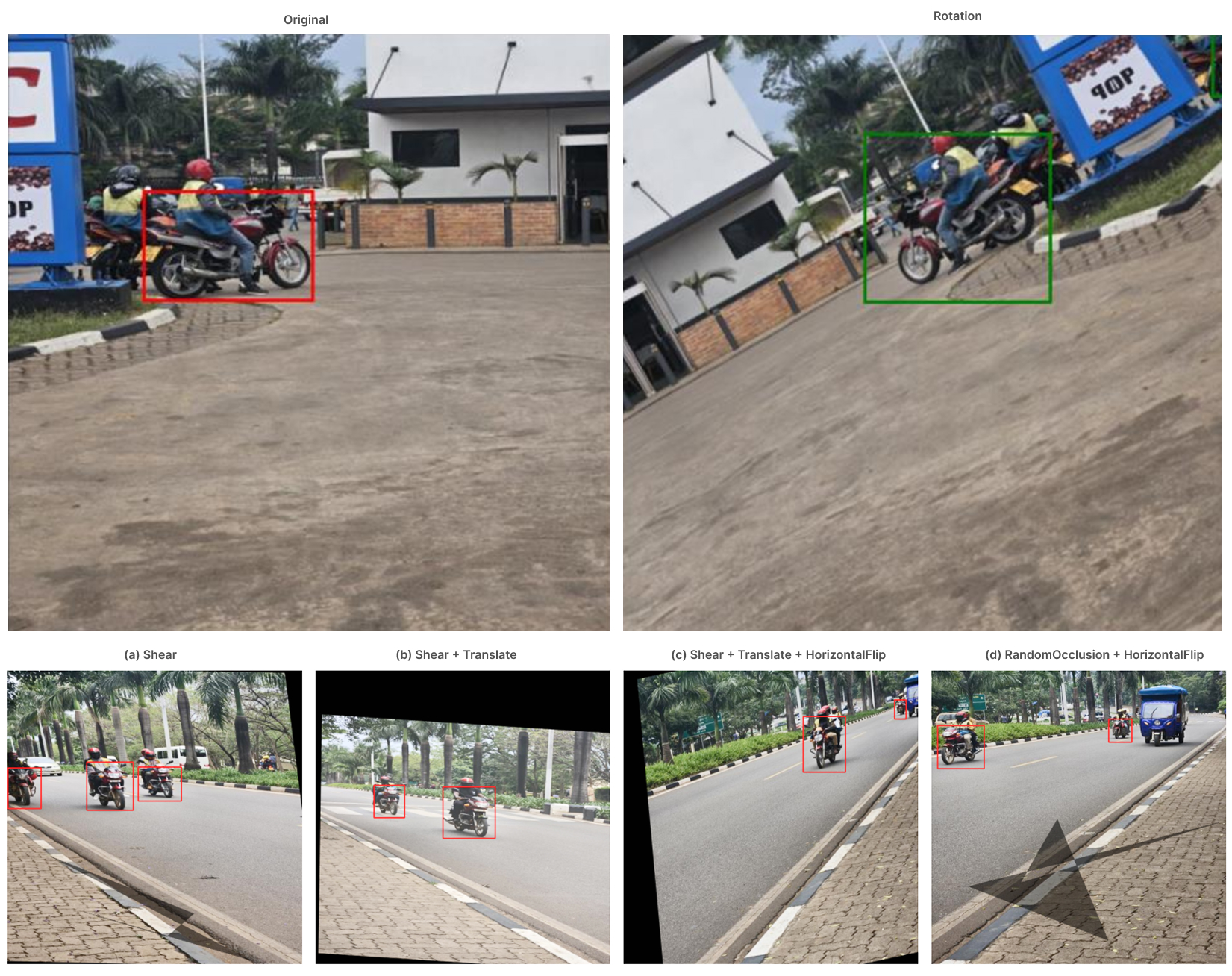}
    \caption{Example augmentations applied to a motorbike image from the Kigali dataset: (a) HorizontalFlip, (b) RandomBrightnessContrast, (c) ShiftScaleRotate, (d) CoarseDropout.}
    \label{fig:augmentations}
\end{figure}

To enhance robustness, augmentations were applied, tailored to each model. YOLOv5 used HorizontalFlip (p=0.7) and ColorJitter (brightness=0.2, contrast=0.2, saturation=0.2). Faster R-CNN applied HorizontalFlip (p=0.7) and ColorJitter. SSD employed HorizontalFlip (p=0.5), RandomBrightnessContrast (p=0.2), ColorJitter, and GaussianNoise (variance 5–20). RetinaNet used HorizontalFlip (p=0.5), RandomBrightnessContrast (p=0.2), RandomRotate90 (p=0.2), ShiftScaleRotate (p=0.2), RGBShift (p=0.2), and CoarseDropout. Figure \ref{fig:augmentations} shows examples of augmentations applied to a motorbike image, with bounding boxes in COCO format.

\subsection{Model Implementations}
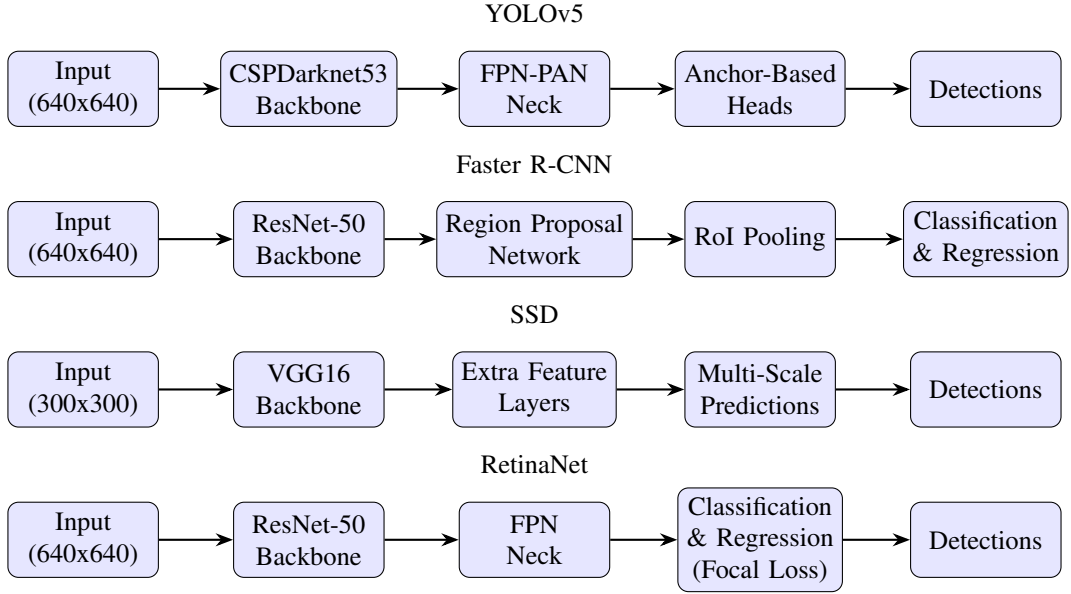
\begin{figure*}[t]
    \centering
    \begin{tikzpicture}[
        box/.style={rectangle, draw, rounded corners, minimum height=1cm, minimum width=2cm, align=center, fill=blue!10},
        arrow/.style={-Stealth, thick},
        label/.style={font=\small, midway, above}
    ]
        \node[box] (yolo_input) at (0, 6) {Input\\(640x640)};
        \node[box] (yolo_backbone) at (3, 6) {CSPDarknet53\\Backbone};
        \node[box] (yolo_neck) at (6, 6) {FPN-PAN\\Neck};
        \node[box] (yolo_head) at (9, 6) {Anchor-Based\\Heads};
        \node[box] (yolo_output) at (12, 6) {Detections};
        \draw[arrow] (yolo_input) -- (yolo_backbone);
        \draw[arrow] (yolo_backbone) -- (yolo_neck);
        \draw[arrow] (yolo_neck) -- (yolo_head);
        \draw[arrow] (yolo_head) -- (yolo_output);
        \node at (6, 7) {YOLOv5};

        \node[box] (frcnn_input) at (0, 4) {Input\\(640x640)};
        \node[box] (frcnn_backbone) at (3, 4) {ResNet-50\\Backbone};
        \node[box] (frcnn_rpn) at (6, 4) {Region Proposal\\Network};
        \node[box] (frcnn_roi) at (9, 4) {RoI Pooling};
        \node[box] (frcnn_head) at (12, 4) {Classification\\\& Regression};
        \draw[arrow] (frcnn_input) -- (frcnn_backbone);
        \draw[arrow] (frcnn_backbone) -- (frcnn_rpn);
        \draw[arrow] (frcnn_rpn) -- (frcnn_roi);
        \draw[arrow] (frcnn_roi) -- (frcnn_head);
        \node at (6, 5) {Faster R-CNN};

        \node[box] (ssd_input) at (0, 2) {Input\\(300x300)};
        \node[box] (ssd_backbone) at (3, 2) {VGG16\\Backbone};
        \node[box] (ssd_layers) at (6, 2) {Extra Feature\\Layers};
        \node[box] (ssd_head) at (9, 2) {Multi-Scale\\Predictions};
        \node[box] (ssd_output) at (12, 2) {Detections};
        \draw[arrow] (ssd_input) -- (ssd_backbone);
        \draw[arrow] (ssd_backbone) -- (ssd_layers);
        \draw[arrow] (ssd_layers) -- (ssd_head);
        \draw[arrow] (ssd_head) -- (ssd_output);
        \node at (6, 3) {SSD};

        \node[box] (retina_input) at (0, 0) {Input\\(640x640)};
        \node[box] (retina_backbone) at (3, 0) {ResNet-50\\Backbone};
        \node[box] (retina_fpn) at (6, 0) {FPN\\Neck};
        \node[box] (retina_head) at (9, 0) {Classification\\\& Regression\\(Focal Loss)};
        \node[box] (retina_output) at (12, 0) {Detections};
        \draw[arrow] (retina_input) -- (retina_backbone);
        \draw[arrow] (retina_backbone) -- (retina_fpn);
        \draw[arrow] (retina_fpn) -- (retina_head);
        \draw[arrow] (retina_head) -- (retina_output);
        \node at (6, 1) {RetinaNet};
    \end{tikzpicture}
    \caption{High-level architectures of YOLOv5, Faster R-CNN, SSD, and RetinaNet.}
    \label{fig:architectures}
\end{figure*}

All models were implemented in PyTorch using torchvision implementations with transfer learning, leveraging pretrained weights to adapt to the Kigali dataset. Figure \ref{fig:architectures} illustrates their architectures.

\subsubsection{YOLOv5}
The official YOLOv5s model \cite{jocher2020yolov5}, implemented via torchvision, used COCO-pretrained weights (\texttt{yolov5s.pt}). A custom implementation was attempted, replicating the CSPDarknet53 backbone, FPN-PAN neck, and anchor-based heads, but failed (mAP 0.0012) due to limited training (5 epochs), as discussed in Section \ref{sec:analysis}. The official model was fine-tuned for 100 epochs on dual T4 GPUs.

\subsubsection{Faster R-CNN}
Faster R-CNN \cite{ren2015faster}, implemented via torchvision with COCO-pretrained ResNet-50-FPN weights, was fine-tuned for 55 epochs with a warmup strategy.

\subsubsection{SSD}
SSD300 \cite{liu2016ssd}, implemented via torchvision with ImageNet-pretrained VGG16 weights, was modified for 2-class predictions and fine-tuned for 10 epochs.

\subsubsection{RetinaNet}
RetinaNet \cite{lin2017focal}, implemented via torchvision with COCO-pretrained ResNet-50-FPN weights, was fine-tuned for 22 epochs with updated heads for 2 classes.

\subsection{Hyperparameters}
\begin{table*}[t]
    \centering
    \caption{Hyperparameter Settings}
    \label{tab:hyperparameters}
    \begin{tabularx}{\textwidth}{XXXXX}
        \toprule
        \textbf{Model} & \textbf{Optimizer} & \textbf{Learning Rate} & \textbf{Batch Size} & \textbf{Epochs} \\
        \midrule
        YOLOv5 & SGD & 0.01 & 32 & 100 \\
        Faster R-CNN & Adam & 0.001 (warmup) & 4 & 55 \\
        SSD & Adam & 0.001 (reduced to 2e-5) & 4 & 10 \\
        RetinaNet & AdamW & 0.0001 & 4 & 22 \\
        \bottomrule
    \end{tabularx}
\end{table*}

\subsection{Evaluation Metrics}
Models were evaluated on the test set using IoU, precision, recall, mAP@0.5, mAP@0.5:0.95, and FPS. SSD was analyzed at confidence thresholds (0.1–0.7).

\section{Results}
\begin{table*}[t]
    \centering
    \caption{Performance Comparison on Test Set}
    \label{tab:results}
    \begin{tabularx}{\textwidth}{XXXXXX}
        \toprule
        \textbf{Model} & \textbf{mAP@0.5:0.95} & \textbf{mAP@0.5} & \textbf{Precision} & \textbf{Recall} & \textbf{FPS} \\
        \midrule
        YOLOv5 & 0.5223 & 0.8511 & 0.6702 & 0.8511 & 24.7 \\
        Faster R-CNN & 0.3320 & 0.7592 & 0.8837 & 0.7755 & 9.27 \\
        SSD & 0.0776 & 0.4200 & 0.3085 & 0.1033 & 49.67 \\
        RetinaNet & 0.6013 & 0.8020 & 0.7500 & 0.6333 & 8.79 \\
        \bottomrule
    \end{tabularx}
\end{table*}
Table \ref{tab:results} presents performance metrics, highlighting trade-offs between accuracy and speed.

\section{Analysis}
\label{sec:analysis}
\subsection{YOLOv5}
The official YOLOv5s model achieved high mAP@0.5:0.95 (0.5223) and FPS (24.7), suitable for real-time navigation. A custom implementation failed (mAP 0.0012) due to target encoding errors, loss instability, anchor misalignment, and insufficient training (5 epochs), highlighting challenges in replicating complex architectures.

\subsection{Faster R-CNN}
Faster R-CNN’s high precision (0.8837) and moderate mAP (0.3320) reflect strong localization, but slow FPS (9.27) limits real-time use. It struggled with small motorbikes due to dataset bias.

\subsection{SSD}
SSD’s high FPS (49.67) suits edge devices, but low mAP@0.5:0.95 (0.0776) versus mAP@0.5 (0.4200) indicates poor localization precision, typical of its simpler feature extraction \cite{liu2016ssd}. It missed small or occluded motorbikes, common in Kigali.

\subsection{RetinaNet}
RetinaNet’s high mAP@0.5:0.95 (0.6013) and IoU (0.8122) reflect robust localization, aided by Focal Loss addressing class imbalance. However, it struggled with small objects, and slow FPS (8.79) limits real-time use.

\section{Discussion}
\label{sec:discussion}
Kigali’s non-compliant motorcycle taxis create unpredictable traffic, challenging autonomous systems. The small dataset (198 images, 30 test annotations) limits statistical significance and generalizability of mAP values, as a single outlier image can skew results. This constraint particularly affected SSD and the custom YOLOv5 implementation, while RetinaNet’s Focal Loss mitigated some imbalance issues but was limited by dataset bias toward larger motorbikes. YOLOv5’s balance of speed and accuracy makes it ideal for real-time navigation, while RetinaNet’s localization is safety-critical. Faster R-CNN’s precision is offset by slow inference, and SSD’s speed is marred by low accuracy.

\subsection{Error Analysis}
YOLOv5 occasionally missed small or distant motorbikes due to anchor misalignment. Faster R-CNN excelled in dense scenes but struggled with small objects. SSD frequently failed on occluded motorbikes due to weak semantic features \cite{arxiv2017fssd}. RetinaNet handled occlusions better but missed small objects, reflecting dataset limitations.

\subsection{Proposed Improvements}
For future work, we recommend:
\begin{itemize}
    \item \textbf{Dataset Expansion}: Collect a larger, more diverse dataset (e.g., 1000+ images) to improve generalizability and reduce bias toward larger motorbikes.
    \item \textbf{YOLOv5-Simple}: Linear box prediction, capped loss, anchor-free heads.
    \item \textbf{Faster R-CNN}: Enhanced augmentations (e.g., rotations, blur).
    \item \textbf{SSD}: Feature fusion modules (FSSD), lower threshold (0.3), higher NMS IoU (0.7), 50 epochs.
    \item \textbf{RetinaNet}: Diverse samples, adjusted anchor scales.
    \item \textbf{Edge Optimization}: Quantization to 8-bit integers and lightweight backbones (e.g., MobileNetV2) for all models \cite{beei2025}.
\end{itemize}

\section{Conclusion}
YOLOv5’s high mAP (0.5223) and FPS (24.7) make it suitable for Kigali’s real-time navigation. RetinaNet’s localization (mAP 0.6013, IoU 0.8122) is safety-critical, Faster R-CNN offers high precision (0.8837), and SSD prioritizes speed (49.67 FPS). Dataset limitations and model complexities highlight the need for larger datasets and optimized architectures to ensure safe autonomous navigation in Kigali.

\bibliographystyle{IEEEtran}

\end{document}